\pdfoutput=1

\documentclass[11pt]{article}

\usepackage[final]{acl}

\usepackage{times}
\usepackage{latexsym}

\usepackage[T1]{fontenc}

\usepackage[utf8]{inputenc}

\usepackage{microtype}

\usepackage{inconsolata}

\usepackage{mathtools}

\usepackage{amsmath}
\usepackage{amsfonts} 
\usepackage{lipsum} 
\usepackage{algorithm}
\usepackage{algorithmic}

\usepackage{graphicx}
\usepackage{transparent}
\usepackage{soul}
\usepackage{calc}
\definecolor{darkred}{HTML}{bb0000}

\usepackage{caption} 
\usepackage{courier}

\usepackage{booktabs}
\usepackage{multirow}
\usepackage{makecell}
\usepackage{arydshln}
\usepackage{color}
\usepackage{colortbl}

\usepackage{subcaption}
\usepackage{amssymb}
\usepackage{gensymb}

\usepackage{enumitem}

\usepackage{nccmath}
\usepackage{hyperref}
\usepackage{grffile}
\usepackage{tabularx}
\usepackage{nicefrac}       
\usepackage{adjustbox}

\usepackage{wrapfig}
\usepackage{hhline}

\usepackage{pifont}

\usepackage{xspace}

\usepackage{wasysym}

\newcommand\blfootnote[1]{%
  \begingroup
  \renewcommand\thefootnote{}\footnote{#1}%
  \addtocounter{footnote}{-1}%
  \endgroup
}

\title{ChatLang-8: An LLM-Based Synthetic Data Generation Framework for Grammatical Error Correction}

\author{Jeiyoon Park$^{1}$,\;\;\; Chanjun Park$^{2\dagger}$, \;\;\; Heuiseok Lim$^{3\dagger}$
\\
  $^{1}$ Atommerce,\;\; $^{2}$ Upstage AI,\;\; $^{3}$ Korea University  \\
  \texttt{jypark1@atommerce.com} \\
  \texttt{chanjun.park@upstage.ai} \\
  \texttt{limhseok@korea.ac.kr}
}

\begin{document}
\maketitle
\begin{abstract}
\blfootnote{$^\dagger$ Corresponding Author}
We explore and improve the capabilities of LLMs to generate data for grammatical error correction (GEC). When merely producing parallel sentences, their patterns are too simplistic to be valuable as a corpus. To address this issue, we propose an automated framework that includes a Subject Selector, Grammar Selector, Prompt Manager, and Evaluator. Additionally, we introduce a new dataset for GEC tasks, named \textbf{ChatLang-8}, which encompasses eight types of subject nouns and 23 types of grammar. It consists of 1 million pairs featuring human-like grammatical errors. Our experiments reveal that ChatLang-8 exhibits a more uniform pattern composition compared to existing GEC datasets. Furthermore, we observe improved model performance when using ChatLang-8 instead of existing GEC datasets. The experimental results suggest that our framework and ChatLang-8 are valuable resources for enhancing ChatGPT's data generation capabilities.

\end{abstract}

\section{Introduction}

With the advent of large language models (LLM) \cite{openai2023gpt4,OpenAI_ChatGPT,openai2023assistants}, fundamental paradigm shift is underway and many subfields of NLP such as machine translation \cite{jiao2023chatgpt}, summarization \cite{yang2023exploring}, question answering \cite{omar2023chatgpt}, and grammatical error correction \cite{wu2023chatgpt,fang2023chatgpt} are being validated by LLM \cite{bang2023multitask}. However, there still lacks investigations into the ability of LLM to correct grammatical errors in text. For instance, ChatGPT gets much lower automatic evaluation scores \cite{wu2023chatgpt} than GECToR \cite{omelianchuk-etal-2020-gector}, based on pretrained RoBERTa model, and GEC product \cite{Grammarly_User}. ChatGPT also suffers from \textit{Overcorrection} problem \cite{fang2023chatgpt} and shows much lower performance than TagGEC \cite{stahlberg-kumar-2021-synthetic}, T5 large, and T5 xxl \cite{rothe-2021-simple}, despite its cutting-edge prompting techniques.

In order to solve these problems, it is indispensable to improve the performance of existing methods or leverage a hybrid model consisting of a GEC model and LLM, using high-quality training datasets. However, these approaches require a lot of manually annotated sentence pairs and these pairs are very expensive and even hard to obtain. In addition, LLM's data generation capabilities are not sufficiently corroborated and reinforced yet. 

\begin{table}[t!]
\small
\centering
\resizebox{0.46\textwidth}{!}{
\begin{tabular}{p{1.9cm} | p{6cm}}
\toprule
 \multicolumn{1}{c|}{\textbf{Method}} &  \multicolumn{1}{c}{\textbf{Generated Subjects}} \\
\cmidrule(lr){1-2}
\multirow{5}{*}{\textbf{Prompt (\textcolor[RGB]{128,0,0}{\ding{54}} SS)}} &\{\textbf{'I'}: 42, \textbf{'She'}: 22, \textbf{'He'}: 11, \textbf{'Dog'}: 5, \textbf{'Cat'}: 4, \textbf{'Children'}: 2, \textbf{'Boys'}: 1, \textbf{'Dogs'}: 1, \textbf{'Book'}: 1, \textbf{'They'}: 1, \textbf{'Person'}: 1, \textbf{'Child'}: 1, \textbf{'You'}: 1, \textbf{'It'}: 1, \textbf{'House'}: 1, \textbf{'John and Sarah'}: 1, \textbf{'Restaurant'}: 1, \textbf{'Chair'}: 1, \textbf{'Elephant'}: 1, \textbf{'Friend'}: 1\}\\ 
\cmidrule(lr){1-2}
\multirow{21}{*}{\textbf{Prompt (\textcolor[RGB]{0,128,0}{\ding{57}} SS)}} &\{\textbf{'I'}: 6, \textbf{'Book'}: 4, \textbf{'Patience'}: 4, \textbf{'Computer'}: 3, \textbf{'Water'}: 3, \textbf{'Monkeys'}: 3, \textbf{'Dog'}: 3, \textbf{'Headphone'}: 2, \textbf{'Tree'}: 2, \textbf{'Courage'}: 2, \textbf{'Flock'}: 2, \textbf{'Cat'}: 2, \textbf{'Car and Motocycle'}: 2, \textbf{'Firefighter'}: 2, \textbf{'Chair'}: 2, \textbf{'Everest'}: 2, \textbf{'Jellyfish'}: 2, \textbf{'Dogs'}: 2, \textbf{'Eiffel Tower'}: 1, \textbf{'Cups'}: 1, \textbf{'Wheelchair'}: 1, \textbf{'Compassion'}: 1, \textbf{'Guitar'}: 1, \textbf{'Freedom'}: 1, \textbf{'Perserverance'}: 1, \textbf{'Grace'}: 1, \textbf{'Camera'}: 1, \textbf{'She'}: 1, \textbf{'Authenticity'}: 1, \textbf{'Taj Mahal'}: 1, \textbf{'Elephants'}: 1, \textbf{'Children'}: 1, \textbf{'Windmill'}: 1, \textbf{'Road'}: 1, \textbf{'Shoe'}: 1, \textbf{'Happiness'}: 1, \textbf{'Refrigerator and Oven'}: 1, \textbf{'Pillow'}: 1, \textbf{'Soldiers'}: 1, \textbf{'Cupcake'}: 1, \textbf{'Airplane'}: 1, \textbf{'Chairs'}: 1, \textbf{'Bird'}: 1, \textbf{'Spoon'}: 1, \textbf{'Barack Obama'}: 1, \textbf{'Serenity and peace'}: 1, \textbf{'Great Barrier'}: 1, \textbf{'Thunderstorm'}: 1, \textbf{'Door'}: 1, \textbf{'Leaves'}: 1, \textbf{'Humility'}: 1, \textbf{'Mona Lisa'}: 1, \textbf{'Desk'}: 1, \textbf{'Snowman and Snowwoman'}: 1, \textbf{'John'}: 1, \textbf{'Pizza'}: 1, \textbf{'Place'}: 1, \textbf{'Herd'}: 1, \textbf{'Cows'}: 1, \textbf{'Bookshelf'}: 1, \textbf{'Toothbrush'}: 1, \textbf{'Respect'}: 1, \textbf{'Diversity'}: 1, \textbf{'Ink'}: 1, \textbf{'Beatles'}: 1, \textbf{'Plant'}: 1, \textbf{'Pride'}: 1, \textbf{'Wolves'}: 1, \textbf{'Congress'}: 1, \textbf{'Honesty'}: 1\}\\
\bottomrule
\end{tabular}}
\linespread{1}
\caption{100 sentences were randomly sampled from each generated result.}
\label{Tab:Generated_subjects}
\end{table}

\textbf{Contributions.} In this paper, (i) we investigate LLM's ability to generate synthetic data, (ii) propose an automatic framework to bolster up the quality of generated data, and (iii) leverage a new GEC dataset, ChatLang-8, which is composed of 1M pairs with a distribution of human-like grammatical errors. We find that ChatLang-8 has a more uniform pattern composition than the existing GEC datasets. We also observe that the model performed better when trained with ChatLang-8 than with the existing GEC dataset with the same corpus size. To the best of our knowledge, we are the first to introduce a GEC dataset using LLM and to demonstrate it outperforms previous human-generated corpora.

\section{ChatLang-8}

The foremost goal is to generate data containing human-like grammatical mistakes. Thus, we aim to create parallel sentence pairs with a distribution of human-like grammatical errors, not unrealistic pairs \cite{grundkiewicz2014the}. When generating pairs, one of the most intuitive ways is to control all elements of the statement. However, because of productivity of language \cite{o2011productivity}, sentences can be infinitely long with recursive rules. It is impossible to control all elements of the generated pairs, and even this approach rather deteriorates the quality of data. Therefore, we first hypothesize that the main factors determining the quality of pairs are the subject types and grammatical error types, and then we diversify them based on our framework.

\begin{table}[t]
\small
\centering
\resizebox{0.46\textwidth}{!}{
\begin{tabular}{p{1.9cm} | p{6cm}}
\toprule
 \multicolumn{1}{c|}{\textbf{Method}} &  \multicolumn{1}{c}{\textbf{Generated Conjunction Errors}} \\
\cmidrule(lr){1-2}
\multirow{1}{*}{\textbf{Prompt (\textcolor[RGB]{128,0,0}{\ding{54}} GS)}} &\{\textbf{'And'}: 59, \textbf{'But'}: 38, \textbf{'Nor'}: 2, \textbf{'Or'}: 1\}\\ 
\cmidrule(lr){1-2}
\multirow{20}{*}{\textbf{Prompt (\textcolor[RGB]{0,128,0}{\ding{57}} GS)}} &\{\textbf{'The...the'}: 6, \textbf{'Neither...nor'}: 6, \textbf{'Both…and'}: 4, \textbf{'So'}: 4, \textbf{'While'}: 4, \textbf{'When'}: 3, \textbf{'So...that:'}: 3, \textbf{'Yet'}: 3, \textbf{'Rather...than'}: 3, \textbf{'Now that'}: 3, \textbf{'And'}: 3, \textbf{'But'}: 3, \textbf{'As'}: 3, \textbf{'Nor'}: 2, \textbf{'Before'}: 2, \textbf{'As soon as'}: 2, \textbf{'As long as'}: 2, \textbf{'As well as'}: 2, \textbf{'Either...or'}: 2, \textbf{'Even if'}: 2, \textbf{'If'}: 2, \textbf{'Or'}: 2, \textbf{'Although'}: 2, \textbf{'Hardly...when'}: 1, \textbf{'Because'}: 1, \textbf{'In case'}: 1, \textbf{'Once'}: 1, \textbf{'Not...but'}: 1, \textbf{'Besides'}: 1, \textbf{'Since'}: 1, \textbf{'Whether...or'}: 1, \textbf{'If...then'}: 1, \textbf{'As many...as'}: 1, \textbf{'So that'}: 1, \textbf{'Not so...as'}: 1, \textbf{'Supposing'}: 1, \textbf{'Not only...but also'}: 1, \textbf{'Provided that'}: 1, \textbf{'For'}: 1, \textbf{'Till'}: 1, \textbf{'Even though'}: 1, \textbf{'Too...to'}: 1, \textbf{'Nonetheless'}: 1, \textbf{'As...as'}: 1, \textbf{'By the time'}: 1, \textbf{'Still'}: 1, \textbf{'Even if'}: 1, \textbf{'No sooner...than'}: 1, \textbf{'That'}: 1, \textbf{'Scarcely...when'}: 1, \textbf{'In order to'}: 1, \textbf{'If only'}: 1, \textbf{'Indeed'}: 1, \textbf{'Otherwise'}: 1, \textbf{'Though'}: 1\}\\
\bottomrule
\end{tabular}}
\linespread{1}
\caption{100 sentences were randomly sampled from each generated result.}
\label{Tab:Generated_grammar_error}
\end{table}

\subsection{Subject Selector}
\label{Sec:2.1}

One naive approach is to design the prompt to generate pairs without considering the subject element. However, as shown in Table \ref{Tab:Generated_subjects}, when simply generating parallel sentences, the type of subject of the sentence becomes too limited. To mitigate this risk, we propose \textit{Subject Selector (SS)}, a novel algorithm for diversifying subject $S$ of parallel sentences. First, we predefine the type of subject $S_k$ to be created\footnote{Predefined types: Common Noun, Proper Noun, Collective Noun, Compound Noun, Concrete Noun, Abstract Noun, Countable Noun, and Uncountable Noun.}. Then Subject Selector randomly choose a type of subject $S_k$ among the predefined noun types and generate subject candidates as large as the window size. For example, if the noun type selected by Subject Selector is \textit{proper noun} and the window size is 30, 30 proper nouns are created. Then, one of the candidates is randomly determine as a subject of a sentence.

\subsection{Grammar Selector}
\label{Sec:2.2}

Similar to the problem in Section \ref{Sec:2.1}, a vanilla prompt to generate pairs of wrong and correct sentences produces an extremely constrained set of sentences. Table \ref{Tab:Generated_grammar_error} shows that the method just specifying the grammar type also could not ameliorate this problem. For instance, when generating data related to \textit{Conjunction}, most of the outputs consist of \textit{And}, and \textit{But}. In this paper, we propose \textit{Grammar Selector (GS)} to consider various grammar errors $G$ and thereby improve the quality of generated data. We take into account the same grammatical error categories $G_l$ in \cite{bryant-etal-2017-automatic, bryant-etal-2019-bea}. Unlike Subject Selector in Section \ref{Sec:2.1}, Grammar Selector first divides the grammar errors into two types: Errors that require diversification of grammar types and those that do not. For example, \textit{Word Order (WO)} and \textit{Spelling (SPELL)} don't need to consider patterns as large as the window size: We can get good results just by instructing the LLM to create representative of real human errors.



\subsection{Prompt Manager}
\label{Sec:2.3}

Given a subject and grammatical error generated from Section \ref{Sec:2.1} and Section \ref{Sec:2.2} respectively, \textit{Prompt Manager} combines them into one prompt. To improve production stability, we incorporate the CoT technique \cite{Wei2022ChainOT} into the Prompt Manager. To circumvent unnecessary errors, Prompt Manager matches a subject of wrong sentence to a subject of right sentence. Prompt Manager maintains the rest of the elements are the same, except for the grammatical error of parallel sentences.

\subsection{Evaluator}
\label{Sec:2.4}

Although generated pairs can be multifarious by exploiting Subject Selector, Grammar Selector, and Prompt Manager, due to uniform randomness of our method to encourage diversity, data that violates prompt instructions were generated occasionally. For example, a prompt \textit{subject: Concrete Noun (car), and grammar\_ type: Verb (swim)} created pairs \textit{"Wrong: The car swims to the shore.", "Right: The car drives to the shore."}. They are the result of breaking the instruction to use the swim verb correctly (i.e., sentences related to verb tense or active/passive voice errors should have been output, not a semantic error.). To alleviate this risk, we propose \textit{Evaluator}, a simple LLM-based evaluator using four criteria $C$ (Figure \ref{Fig:Evaluator}). Evaluator $\mathcal{E}$ evaluates itself for compliance with the data creation process and employs a unanimous vote: If even a single criterion $C_i$ is not met, the generated result is discarded:  

\begin{figure}[t!]
    \centering 
    \includegraphics[width=1\linewidth]{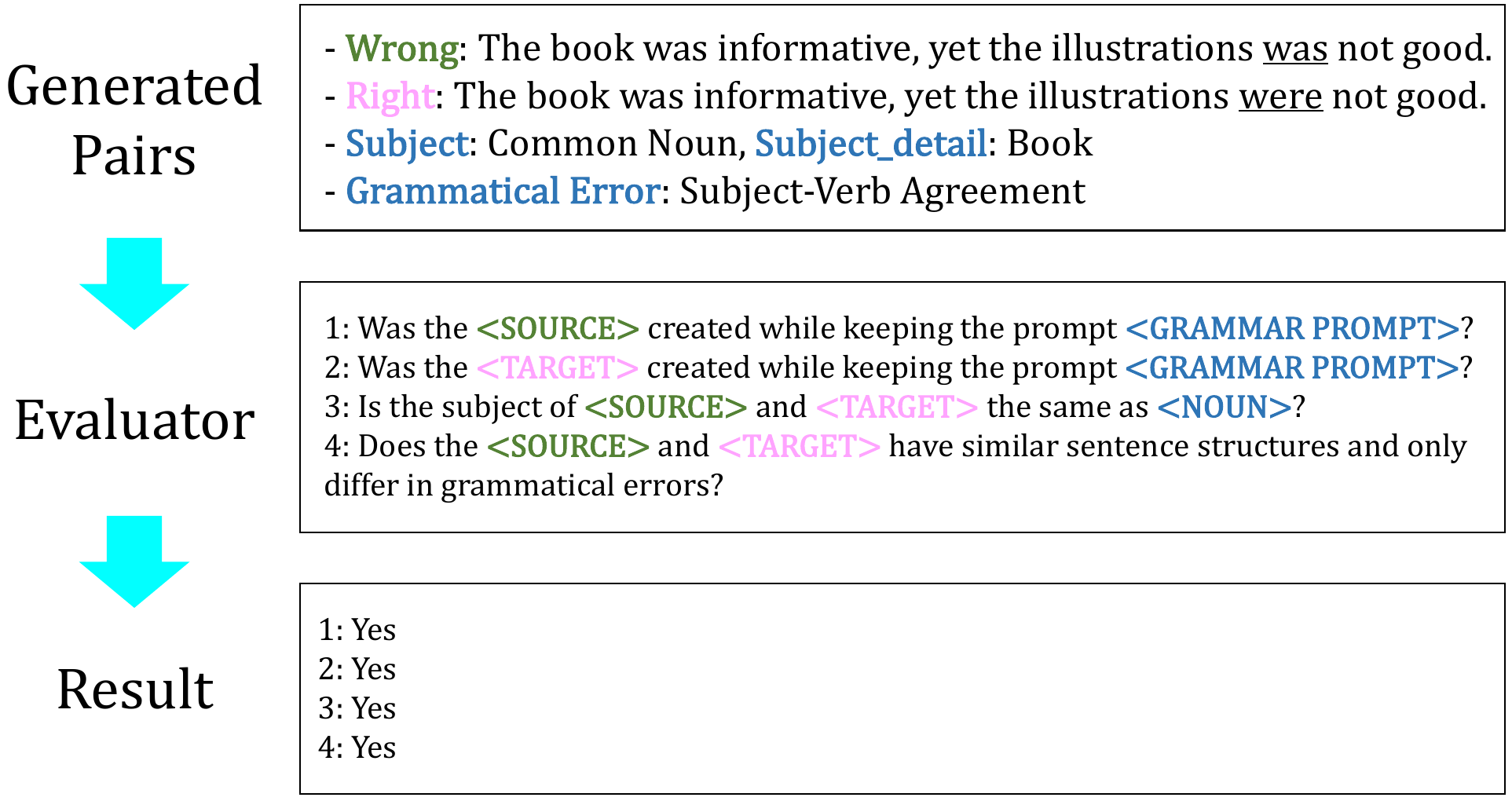}
    \caption{The pipeline of Evaluator.}
    \label{Fig:Evaluator}
    \vspace{-1em}
\end{figure}

\begin{equation}
\mathcal{E}(D)=\left\{
\begin{array}{ll}
\displaystyle{1}, \quad if \quad C_1 \wedge C_2 \wedge C_3 \wedge C_4\\[4mm]
\displaystyle{0}, \quad otherwise
\end{array}\right.
\end{equation}

\subsection{Dataset Details}
\label{Sec:2.5}
Our implementation utilizes GPT-3.5 Turbo. We gathered approximately 1M pairs that cover eight subject type and 23 grammar type. Table \ref{Tab:Corpus} shows statistics of \textit{ChatLang-8} (ours), FCE \cite{yannakoudakis-etal-2011-new}, Lang-8 \cite{mizumoto-etal-2012-effect,tajiri-etal-2012-tense}, NUCLE \cite{dahlmeier-etal-2013-building} and W\&I+LOCNESS \cite{bryant-etal-2019-bea}. Note that we leverage ChatLang-8 that has a corpus size similar to Lang-8, one of the largest GEC corpus, and much larger than the other corpora. 

\section{Experiments}


\begin{figure}[t]
    \centering 
    \includegraphics[width=1\linewidth]{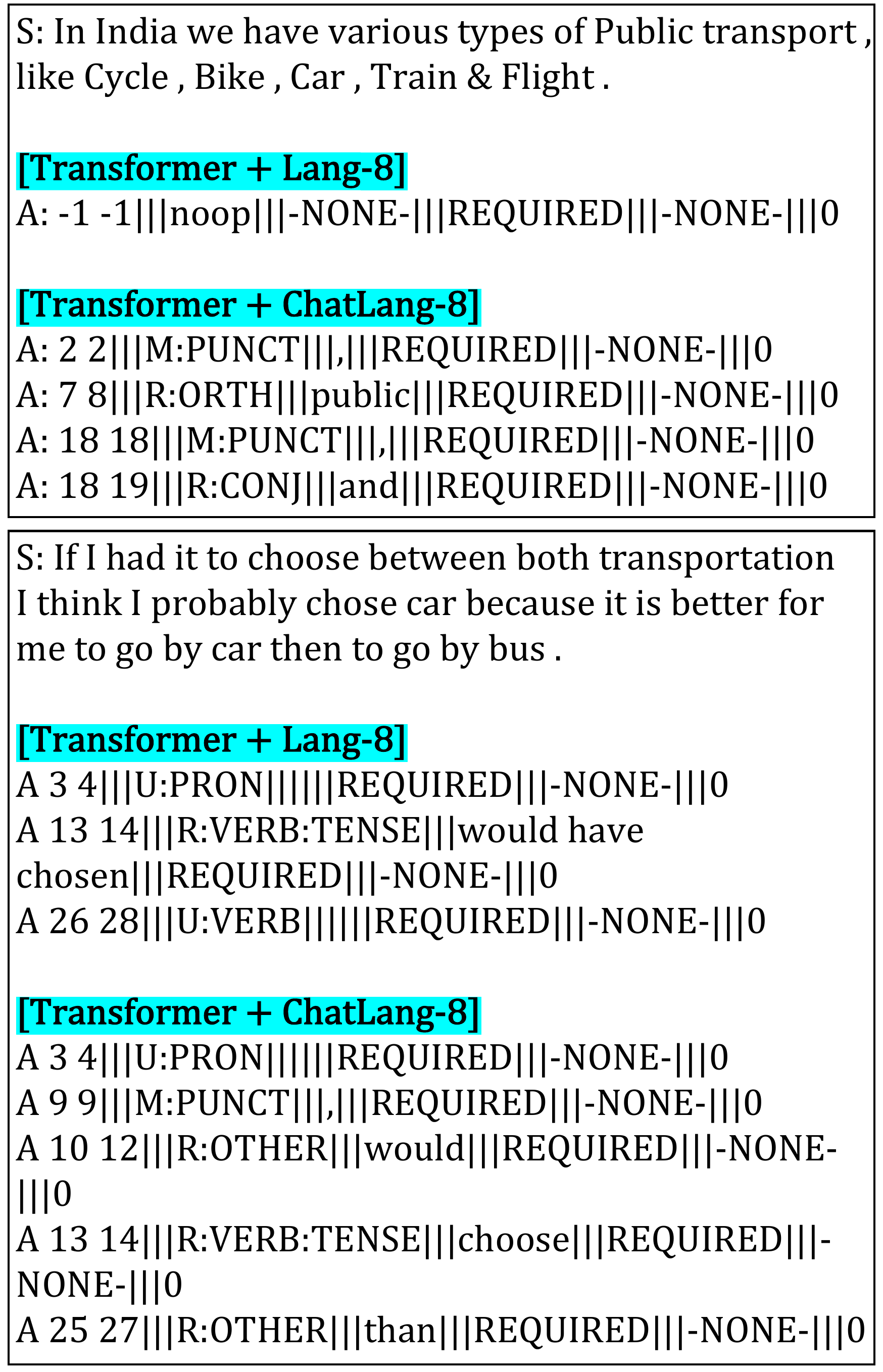}
    \caption{Comparison of M$^2$ outputs of GEC models, trained on Lang-8 and ChatLang-8 respectively.}
    \label{Fig:Qualitative_result}
    \vspace{-1em}
\end{figure}

\subsection{Statistical Analysis}



\begin{figure}[t!]
    \centering 
    \includegraphics[width=1\linewidth]{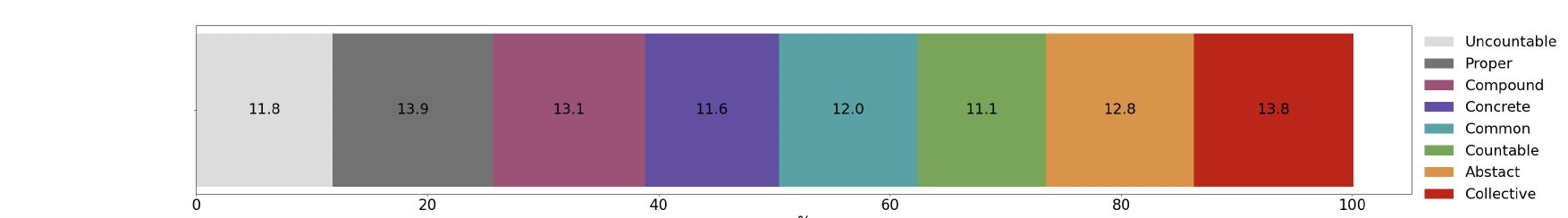}
    \caption{Subject distribution of ChatLang-8.}
    \label{Fig:Subject_bar}
    \vspace{-1em}
\end{figure}

\begin{figure}[t!]
    \centering 
    \includegraphics[width=1\linewidth]{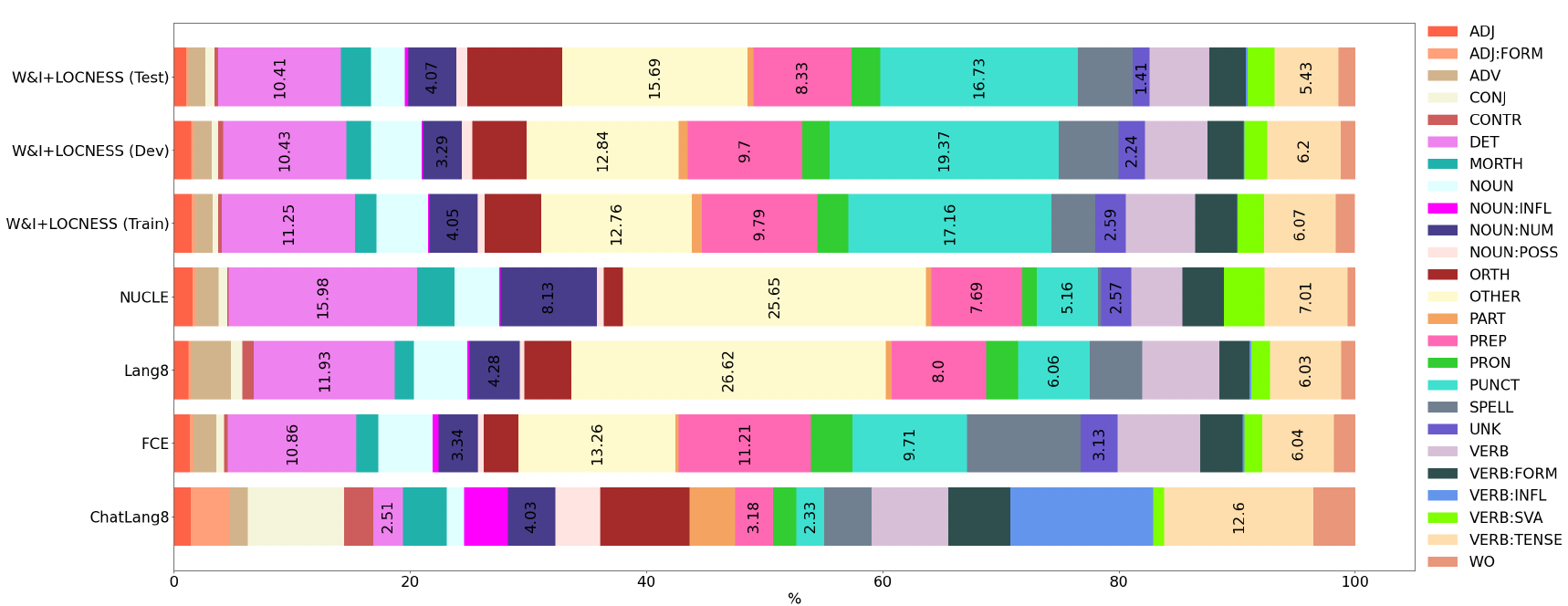}
    \caption{Distributions of the ChatLang-8 and the other datasets with respect to 25 grammatical errors.}
    \label{Fig:Error_distribution}
    \vspace{-1em}
\end{figure}

Note that numerous studies show that the imbalance of a dataset causes performance degradation \cite{huang-etal-2021-balancing,lin2023robust,li-shi-2021-tail}. In order to demonstrate whether the dataset is created evenly, we conduct statistical analysis by comparing ChatLang-8 with existing datasets. Figure \ref{Fig:Subject_bar} shows a subject distribution of ChatLang-8. We observe that the distribution of subject types in ChatLang-8 is evenly distributed. It suggests that the Subject Selector and evaluator affect data quality improvement even though the evaluator abandons all unsuitable data. In Figure \ref{Fig:Error_distribution} and Table \ref{Tab:Error_distribution}, we find that the grammar types of ChatLang-8 are most evenly distributed while the other corpora have highly unbalanced grammar labels. For example, W\&I+LOCNESS dataset has 10.7\% (averaged) of Determiner (DET) errors, and 17.75\% of Punctuation (PUNCT) errors, whereas 0.6133\% of Conjunction (CONJ) and 0.21\% of Adjective Form (ADJ:FORM). Existing datasets have a very small proportion of important grammatical patterns, which can cause performance degradation during model training. In addition, since ChatLang-8 excluded OTHER and UNK labels when generating data, while the other corpora have a great proportion of UNK label (e.g., NUCLE: 25.65\%) and OTHER label (e.g., NUCLE: 2.57\%), it is easier to handle dataset and to predict the model outputs. 

\begin{table}[t]
\centering
\scalebox{0.61}{
\begin{tabular}{lccccc}
\toprule
& ChatLang-8 & FCE & Lang-8 & NUCLE & W\&I+LOCNESS\\
\midrule
Sentences & 1,016,588 & 33,236 & 1,037,561 & 57,151 & 43,169 \\
\bottomrule
\end{tabular}}
\caption{ChatLang-8, FCE, Lang-8, NUCLE, and W\&I+LOCNESS dataset statistics.}
\label{Tab:Corpus}
\end{table}

\begin{table}[!t]
\centering
\scalebox{0.75}{
\begin{tabular}{lcccccc}
    \toprule  
    \multirow{2}{*}{\textbf{CoNLL-2014}\vspace{-2mm}}
    & \multicolumn{3}{c}{\textbf{ChatLang-8}} & \multicolumn{3}{c}{\textbf{Lang-8}}\\
    \cmidrule(lr){2-4} \cmidrule(lr){5-7}
      & \textbf{P} & \textbf{R} & \textbf{F$_\text{0.5}$} & \textbf{P} & \textbf{R} & \textbf{F$_\text{0.5}$}\\
    \midrule  
    Transformer & 47.79 & 26.39 & 41.12 & 54.37 & 13.28 & 33.58 \\
    BART & 52.50 & 34.31 & 47.47 & 27.89 & 16.01 & 24.29 \\
    \bottomrule 
\end{tabular}
}

\begin{center}
    \vspace{1mm}
    \scalebox{0.77}{
    \begin{tabular}{lcccccc}
        \toprule  
        \multirow{2}{*}{\textbf{BEA-2019}\vspace{-2mm}}
        & \multicolumn{3}{c}{\textbf{ChatLang-8}} & \multicolumn{3}{c}{\textbf{Lang-8}}\\
        \cmidrule(lr){2-4} \cmidrule(lr){5-7}
          & \textbf{P} & \textbf{R} & \textbf{F$_\text{0.5}$} & \textbf{P} & \textbf{R} & \textbf{F$_\text{0.5}$}\\
        \midrule  
        Transformer &  34.44 & 19.69 & 29.95 & 44.81 & 8.5 & 24.16\\
        BART &  34.86 & 24.38 & 32.10 & 23.69 & 11.00 & 19.25\\
        \bottomrule 
    \end{tabular}
    }
\end{center}

\begin{center}
    \vspace{1mm}
    \scalebox{0.76}{
    \begin{tabular}{lcccccc}
        \toprule  
        \multirow{2}{*}{\textbf{GMEG-wiki}\vspace{-2mm}}
        & \multicolumn{3}{c}{\textbf{ChatLang-8}} & \multicolumn{3}{c}{\textbf{Lang-8}}\\
        \cmidrule(lr){2-4} \cmidrule(lr){5-7}
          & \textbf{P} & \textbf{R} & \textbf{F$_\text{0.5}$} & \textbf{P} & \textbf{R} & \textbf{F$_\text{0.5}$}\\
        \midrule  
        Transformer &  46.56 & 34.33 & 43.46 & 41.44 & 11.38 & 27.11\\
        BART &  49.09 & 41.77 & 47.42 & 23.54 & 15.68 & 21.39\\
        \bottomrule 
    \end{tabular}
    }
\end{center}
\vspace{-3mm}
\caption{\small
Evaluation results on English CoNLL-2014, BEA-2019, and GMEG-wiki.}
\label{Tab:Results}
\end{table}


\subsection{Quantitative Results} 
\label{Sec:3.2}
To investigate whether ChatLang-8 has value as a dataset, we train a vanilla Transformer \cite{NIPS2017_3f5ee243} and BART \cite{lewis-etal-2020-bart}, much larger model, on both ChatLang-8 and Lang-8 \cite{mizumoto-etal-2012-effect,tajiri-etal-2012-tense}. Note that for the fairness of the experiment, we choose Lang-8, which has a similar dataset structure and corpus size, and which effectively improve performance \cite{lichtarge-etal-2019-corpora,lichtarge-etal-2020-data,flachs-etal-2021-data}. In addition, though there is previous attempt to improve the Lang-8 \cite{flachs-etal-2021-data}, it reports only five error types on BEA test while we considers 25 error types and three benchmarks. Thus, we employ Lang-8 to the experiments, instead of cLang-8. We evaluate performance on three popular GEC benchmarks: CoNLL-2014 \cite{ng-etal-2014-conll}, BEA-2019 \cite{bryant-etal-2019-bea}, and GMEG-wiki \cite{napoles-etal-2019-enabling}. As evaluation metrics, we use M$^2$ scorer \cite{dahlmeier-ng-2012-better} for CoNLL-2014, and ERRANT \cite{bryant-etal-2017-automatic} for BEA-2019 and GMEG-wiki. Table \ref{Tab:Results} shows that GEC models trained on ChatLang-8 can surpass its counterparts which are trained on Lang-8 corpus, by a large margin in both recall (\textbf{R}) and \textbf{F$_{0.5}$}. It represents that the GEC system trained with ChatLang-8 can simultaneously detect multiple types grammatical errors in a sentence and correct them, while system trained with Lang-8 can't. Simply a high \textbf{R} and a low \textbf{F$_{0.5}$} indicate overcorrection. We observe that transformer learned with Lang-8 shows higher precision (\textbf{P}) than its counterpart on CoNLL-2014 and BEA-2019 benchmark evaluations, entailing much lower \textbf{F$_{0.5}$}. This means that it corrects only very obvious grammatical errors and leaves the rest uncorrected. We also find that leveraging a high-quality GEC corpus is overarching problem, as merely enlarging the model does not always ensure improved performance. 

\subsection{Qualitative Results} 
Figure \ref{Fig:Qualitative_result} shows an example and its corresponding M$^2$ outputs, which qualitatively demonstrate the effect of dataset used for training, as shown in Section \ref{Sec:3.2}. We observe that the learned model using ChatLang-8 can detect and correct multiple errors including punctuation (PUNCT), orthography (ORTH), and conjunction (CONJ) while its opponent can not find any errors. Indeed, the output of model learned with ChatLang-8 is \textit{In India, we have various types of public transport, like Cycle, Bike, Car, Train, and Flight}. Results in Figure \ref{Fig:Qualitative_result} demonstrates that ChatLang-8 has the error distribution shown in Table \ref{Tab:Error_distribution} very well, and thereby efficiently helps the learning of the GEC system.  

\section{Conclusion}
We propose a framework with a hypothesize that determining the quality of dataset is the types of subject and grammatical error. We introduce that ChatLang-8, a high-quality dataset for GEC task that is composed of 1M pairs and covers eight subject type and 23 grammar type. We are the first to build a GEC dataset using ChatGPT and verify it outperforms existing corpora. We hope that our method and dataset can help researchers make a better-informed decision about their research.

\clearpage
\section*{Limitations}

\textbf{Factuality and Morality.} Though ChatLang-8 effectively can reflect a distribution of human-like grammatical errors in Figure \ref{Tab:Error_distribution} and guide a GEC model to perform well, due to uniform randomness of our method to encourage diversity, sometimes generated pairs have information that is literally not true. For instance, in Section \ref{Sec:2.1}, Subject Selector, of course, can generate celebrity (e.g., \textit{Michael Jackson} or \textit{Taylor Swift}), or even company name (e.g., \textit{Google} or \textit{Amazon}). By combining the selected subject and grammatical information, sentences that are not true can be generated as follows:

\begin{quote}{\small
\textit{Michael Jackson was known for his music career, in order that he could be more successful, he decided to try his hand at acting.}\\\\
\textit{Taylor Swift is really good at using a computer.}
}
\end{quote}

In addition, the generated sentences may contain biased information such as subjective opinions:

\begin{quote}{\small
\textit{Barack Obama's speeches were always inspiring the crowds.}\\\\
\textit{Google is the best search engine in the world.}
}
\end{quote}

Note that, although most of pairs are normal and fabricated information is found in some generated examples, this factuality problem must be resolved.

On the other hand, we also observed whether ChatLang-8 has a problem with moral deviations. Fortunately, no morality issue was found in most of the pairs constituting ChatLang-8:

\begin{quote}{\small
\textit{The equality between races, genders, and sexual orientations is important.}\\\\
\textit{The importance of maintaining equality among all genders is evident in society.}
}
\end{quote}

However, we found one sentence and removed from ChatLang-8: 

\begin{quote}{\small
\textit{The equality of the sexes is not a big issue.}
}
\end{quote}

These observations suggest that semantic information as well as grammatical information of the generated sentence should be added to the evaluation criteria of our Evaluator.

\textbf{Time and Cost.} As mentioned in Section \ref{Sec:2.5}, Our implementation utilizes gpt3.5-turbo version of ChatGPT. It took about two weeks to gather ChatLang-8. To collect 1M pairs dataset, we spent about 1.1K \textit{dollars}. However, It should be noted that the cost was higher than expected because some pairs are discarded by Evaluator. Evaluator disposed of about 0.7M sentences out of a total of 1.7M generated sentences. We consider that though it is much more efficient than having annotators create the dataset, this a great waste of time and resources. Also, there is a trade-off in our framework: The larger window size, the more diverse and high-quality dataset can be obtained, but the time and cost to create dataset increases accordingly.

\textbf{Discussion and Future Works.} In this paper, we strived to validate the ability of LLM to generate datasets. Of course, efficient dataset acquisition and model training are important challenges in the GEC task, but in fact, we also wanted to know what information is important and should be considered in Natural language generation in order to improve quality of outputs generated by LLM. We hope that this study and its results will be helpful to researchers who have the same concerns as us.

Our direction of future work is threefold:

\begin{enumerate}
    \item We will establish new criteria for factuality and morality, and improve the evaluator through methods such as giving penalties for the generated results that violate the criteria.
    \item We also will devise methods such as soft alignment to minimize the amount of data discarded by Evaluator to prevent resource waste.
    \item Though we conducted an in-depth analysis through statistical, quantitative, and qualitative methods, we relied on LLM for data collection, leaving human evaluation behind. We will experiment with more off-the-shelf baselines by increasing the size of the corpus, and human evaluation will also be conducted on the generated results.
\end{enumerate}



\section*{Ethics Statement}
There are no ethical issues in this study. The data was collected through legitimate methods. Moreover, the experimental results were obtained through an objective comparison. The authors confirm that the research was conducted in accordance with the relevant ethical guidelines and principles.



\bibliography{chatlang8}

\begin{thebibliography}{33}
\expandafter\ifx\csname natexlab\endcsname\relax\def\natexlab#1{#1}\fi

\bibitem[{Bang et~al.(2023)Bang, Cahyawijaya, Lee, Dai, Su, Wilie, Lovenia, Ji, Yu, Chung et~al.}]{bang2023multitask}
Yejin Bang, Samuel Cahyawijaya, Nayeon Lee, Wenliang Dai, Dan Su, Bryan Wilie, Holy Lovenia, Ziwei Ji, Tiezheng Yu, Willy Chung, et~al. 2023.
\newblock A multitask, multilingual, multimodal evaluation of chatgpt on reasoning, hallucination, and interactivity.
\newblock \emph{arXiv preprint arXiv:2302.04023}.

\bibitem[{Bryant et~al.(2019)Bryant, Felice, Andersen, and Briscoe}]{bryant-etal-2019-bea}
Christopher Bryant, Mariano Felice, {\O}istein~E. Andersen, and Ted Briscoe. 2019.
\newblock \href {https://doi.org/10.18653/v1/W19-4406} {The {BEA}-2019 shared task on grammatical error correction}.
\newblock In \emph{Proceedings of the Fourteenth Workshop on Innovative Use of NLP for Building Educational Applications}, pages 52--75, Florence, Italy. Association for Computational Linguistics.

\bibitem[{Bryant et~al.(2017)Bryant, Felice, and Briscoe}]{bryant-etal-2017-automatic}
Christopher Bryant, Mariano Felice, and Ted Briscoe. 2017.
\newblock \href {https://doi.org/10.18653/v1/P17-1074} {Automatic annotation and evaluation of error types for grammatical error correction}.
\newblock In \emph{Proceedings of the 55th Annual Meeting of the Association for Computational Linguistics (Volume 1: Long Papers)}, pages 793--805, Vancouver, Canada. Association for Computational Linguistics.

\bibitem[{Dahlmeier and Ng(2012)}]{dahlmeier-ng-2012-better}
Daniel Dahlmeier and Hwee~Tou Ng. 2012.
\newblock \href {https://aclanthology.org/N12-1067} {Better evaluation for grammatical error correction}.
\newblock In \emph{Proceedings of the 2012 Conference of the North {A}merican Chapter of the Association for Computational Linguistics: Human Language Technologies}, pages 568--572, Montr{\'e}al, Canada. Association for Computational Linguistics.

\bibitem[{Dahlmeier et~al.(2013)Dahlmeier, Ng, and Wu}]{dahlmeier-etal-2013-building}
Daniel Dahlmeier, Hwee~Tou Ng, and Siew~Mei Wu. 2013.
\newblock \href {https://aclanthology.org/W13-1703} {Building a large annotated corpus of learner {E}nglish: The {NUS} corpus of learner {E}nglish}.
\newblock In \emph{Proceedings of the Eighth Workshop on Innovative Use of {NLP} for Building Educational Applications}, pages 22--31, Atlanta, Georgia. Association for Computational Linguistics.

\bibitem[{Fang et~al.(2023)Fang, Yang, Lan, Wong, Hu, Chao, and Zhang}]{fang2023chatgpt}
Tao Fang, Shu Yang, Kaixin Lan, Derek~F Wong, Jinpeng Hu, Lidia~S Chao, and Yue Zhang. 2023.
\newblock Is chatgpt a highly fluent grammatical error correction system? a comprehensive evaluation.
\newblock \emph{arXiv preprint arXiv:2304.01746}.

\bibitem[{Flachs et~al.(2021)Flachs, Stahlberg, and Kumar}]{flachs-etal-2021-data}
Simon Flachs, Felix Stahlberg, and Shankar Kumar. 2021.
\newblock \href {https://aclanthology.org/2021.bea-1.12} {Data strategies for low-resource grammatical error correction}.
\newblock In \emph{Proceedings of the 16th Workshop on Innovative Use of NLP for Building Educational Applications}, pages 117--122, Online. Association for Computational Linguistics.

\bibitem[{Grammarly(2023)}]{Grammarly_User}
Grammarly. 2023.
\newblock \href {https://www.grammarly.com/about} {Grammarly website about us page}.

\bibitem[{Grundkiewicz and Junczys-Dowmunt(2014)}]{grundkiewicz2014the}
Roman Grundkiewicz and Marcin Junczys-Dowmunt. 2014.
\newblock \href {https://www.microsoft.com/en-us/research/publication/wiked-error-corpus-corpus-corrective-wikipedia-edits-application-grammatical-error-correction/} {The wiked error corpus: A corpus of corrective wikipedia edits and its application to grammatical error correction}.
\newblock In \emph{International Conference on Natural Language Processing}, pages 478--490.

\bibitem[{Huang et~al.(2021)Huang, Giledereli, K{\"o}ksal, {\"O}zg{\"u}r, and Ozkirimli}]{huang-etal-2021-balancing}
Yi~Huang, Buse Giledereli, Abdullatif K{\"o}ksal, Arzucan {\"O}zg{\"u}r, and Elif Ozkirimli. 2021.
\newblock \href {https://doi.org/10.18653/v1/2021.emnlp-main.643} {Balancing methods for multi-label text classification with long-tailed class distribution}.
\newblock In \emph{Proceedings of the 2021 Conference on Empirical Methods in Natural Language Processing}, pages 8153--8161, Online and Punta Cana, Dominican Republic. Association for Computational Linguistics.

\bibitem[{Jiao et~al.(2023)Jiao, Wang, Huang, Wang, and Tu}]{jiao2023chatgpt}
Wenxiang Jiao, Wenxuan Wang, JT~Huang, Xing Wang, and ZP~Tu. 2023.
\newblock Is chatgpt a good translator? yes with gpt-4 as the engine.
\newblock \emph{arXiv preprint arXiv:2301.08745}.

\bibitem[{Lewis et~al.(2020)Lewis, Liu, Goyal, Ghazvininejad, Mohamed, Levy, Stoyanov, and Zettlemoyer}]{lewis-etal-2020-bart}
Mike Lewis, Yinhan Liu, Naman Goyal, Marjan Ghazvininejad, Abdelrahman Mohamed, Omer Levy, Veselin Stoyanov, and Luke Zettlemoyer. 2020.
\newblock \href {https://doi.org/10.18653/v1/2020.acl-main.703} {{BART}: Denoising sequence-to-sequence pre-training for natural language generation, translation, and comprehension}.
\newblock In \emph{Proceedings of the 58th Annual Meeting of the Association for Computational Linguistics}, pages 7871--7880, Online. Association for Computational Linguistics.

\bibitem[{Li and Shi(2021)}]{li-shi-2021-tail}
Piji Li and Shuming Shi. 2021.
\newblock \href {https://doi.org/10.18653/v1/2021.acl-long.385} {Tail-to-tail non-autoregressive sequence prediction for {C}hinese grammatical error correction}.
\newblock In \emph{Proceedings of the 59th Annual Meeting of the Association for Computational Linguistics and the 11th International Joint Conference on Natural Language Processing (Volume 1: Long Papers)}, pages 4973--4984, Online. Association for Computational Linguistics.

\bibitem[{Lichtarge et~al.(2020)Lichtarge, Alberti, and Kumar}]{lichtarge-etal-2020-data}
Jared Lichtarge, Chris Alberti, and Shankar Kumar. 2020.
\newblock \href {https://doi.org/10.1162/tacl_a_00336} {Data weighted training strategies for grammatical error correction}.
\newblock \emph{Transactions of the Association for Computational Linguistics}, 8:634--646.

\bibitem[{Lichtarge et~al.(2019)Lichtarge, Alberti, Kumar, Shazeer, Parmar, and Tong}]{lichtarge-etal-2019-corpora}
Jared Lichtarge, Chris Alberti, Shankar Kumar, Noam Shazeer, Niki Parmar, and Simon Tong. 2019.
\newblock \href {https://doi.org/10.18653/v1/N19-1333} {Corpora generation for grammatical error correction}.
\newblock In \emph{Proceedings of the 2019 Conference of the North {A}merican Chapter of the Association for Computational Linguistics: Human Language Technologies, Volume 1 (Long and Short Papers)}, pages 3291--3301, Minneapolis, Minnesota. Association for Computational Linguistics.

\bibitem[{Lin et~al.(2023)Lin, Tan, Nguyen, Lang, Du, Buntine, Beare, Chen, and Gasevic}]{lin2023robust}
Jionghao Lin, Wei Tan, Ngoc~Dang Nguyen, David Lang, Lan Du, Wray Buntine, Richard Beare, Guanliang Chen, and Dragan Gasevic. 2023.
\newblock \href {http://arxiv.org/abs/2304.07499} {Robust educational dialogue act classifiers with low-resource and imbalanced datasets}.

\bibitem[{Mizumoto et~al.(2012)Mizumoto, Hayashibe, Komachi, Nagata, and Matsumoto}]{mizumoto-etal-2012-effect}
Tomoya Mizumoto, Yuta Hayashibe, Mamoru Komachi, Masaaki Nagata, and Yuji Matsumoto. 2012.
\newblock \href {https://aclanthology.org/C12-2084} {The effect of learner corpus size in grammatical error correction of {ESL} writings}.
\newblock In \emph{Proceedings of {COLING} 2012: Posters}, pages 863--872, Mumbai, India. The COLING 2012 Organizing Committee.

\bibitem[{Napoles et~al.(2019)Napoles, N{\u{a}}dejde, and Tetreault}]{napoles-etal-2019-enabling}
Courtney Napoles, Maria N{\u{a}}dejde, and Joel Tetreault. 2019.
\newblock \href {https://doi.org/10.1162/tacl_a_00282} {Enabling robust grammatical error correction in new domains: Data sets, metrics, and analyses}.
\newblock \emph{Transactions of the Association for Computational Linguistics}, 7:551--566.

\bibitem[{Ng et~al.(2014)Ng, Wu, Briscoe, Hadiwinoto, Susanto, and Bryant}]{ng-etal-2014-conll}
Hwee~Tou Ng, Siew~Mei Wu, Ted Briscoe, Christian Hadiwinoto, Raymond~Hendy Susanto, and Christopher Bryant. 2014.
\newblock \href {https://doi.org/10.3115/v1/W14-1701} {The {C}o{NLL}-2014 shared task on grammatical error correction}.
\newblock In \emph{Proceedings of the Eighteenth Conference on Computational Natural Language Learning: Shared Task}, pages 1--14, Baltimore, Maryland. Association for Computational Linguistics.

\bibitem[{O'Donnell et~al.(2011)O'Donnell, Snedeker, Tenenbaum, and Goodman}]{o2011productivity}
Timothy O'Donnell, Jesse Snedeker, Joshua Tenenbaum, and Noah Goodman. 2011.
\newblock Productivity and reuse in language.
\newblock In \emph{Proceedings of the Annual Meeting of the Cognitive Science Society}, volume~33.

\bibitem[{Omar et~al.(2023)Omar, Mangukiya, Kalnis, and Mansour}]{omar2023chatgpt}
Reham Omar, Omij Mangukiya, Panos Kalnis, and Essam Mansour. 2023.
\newblock Chatgpt versus traditional question answering for knowledge graphs: Current status and future directions towards knowledge graph chatbots.
\newblock \emph{arXiv preprint arXiv:2302.06466}.

\bibitem[{Omelianchuk et~al.(2020)Omelianchuk, Atrasevych, Chernodub, and Skurzhanskyi}]{omelianchuk-etal-2020-gector}
Kostiantyn Omelianchuk, Vitaliy Atrasevych, Artem Chernodub, and Oleksandr Skurzhanskyi. 2020.
\newblock \href {https://doi.org/10.18653/v1/2020.bea-1.16} {{GECT}o{R} {--} grammatical error correction: Tag, not rewrite}.
\newblock In \emph{Proceedings of the Fifteenth Workshop on Innovative Use of NLP for Building Educational Applications}, pages 163--170, Seattle, WA, USA → Online. Association for Computational Linguistics.

\bibitem[{OpenAI(2023{\natexlab{a}})}]{openai2023gpt4}
OpenAI. 2023{\natexlab{a}}.
\newblock \href {http://arxiv.org/abs/2303.08774} {Gpt-4 technical report}.

\bibitem[{OpenAI(2023{\natexlab{b}})}]{openai2023assistants}
OpenAI. 2023{\natexlab{b}}.
\newblock \href {https://platform.openai.com/docs/assistants/overview} {no date provided. introducing assistants api}.

\bibitem[{OpenAI(2023{\natexlab{c}})}]{OpenAI_ChatGPT}
OpenAI. 2023{\natexlab{c}}.
\newblock \href {https://openai.com/blog/chatgpt} {no date provided. introducing chatgpt}.

\bibitem[{Rothe et~al.(2021)Rothe, Mallinson, Malmi, Krause, and Severyn}]{rothe-2021-simple}
Sascha Rothe, Jonathan Mallinson, Eric Malmi, Sebastian Krause, and Aliaksei Severyn. 2021.
\newblock \href {https://doi.org/10.18653/v1/2021.acl-short.89} {A simple recipe for multilingual grammatical error correction}.
\newblock In \emph{Proceedings of the 59th Annual Meeting of the Association for Computational Linguistics and the 11th International Joint Conference on Natural Language Processing (Volume 2: Short Papers)}, pages 702--707, Online. Association for Computational Linguistics.

\bibitem[{Stahlberg and Kumar(2021)}]{stahlberg-kumar-2021-synthetic}
Felix Stahlberg and Shankar Kumar. 2021.
\newblock \href {https://aclanthology.org/2021.bea-1.4} {Synthetic data generation for grammatical error correction with tagged corruption models}.
\newblock In \emph{Proceedings of the 16th Workshop on Innovative Use of NLP for Building Educational Applications}, pages 37--47, Online. Association for Computational Linguistics.

\bibitem[{Tajiri et~al.(2012)Tajiri, Komachi, and Matsumoto}]{tajiri-etal-2012-tense}
Toshikazu Tajiri, Mamoru Komachi, and Yuji Matsumoto. 2012.
\newblock \href {https://aclanthology.org/P12-2039} {Tense and aspect error correction for {ESL} learners using global context}.
\newblock In \emph{Proceedings of the 50th Annual Meeting of the Association for Computational Linguistics (Volume 2: Short Papers)}, pages 198--202, Jeju Island, Korea. Association for Computational Linguistics.

\bibitem[{Vaswani et~al.(2017)Vaswani, Shazeer, Parmar, Uszkoreit, Jones, Gomez, Kaiser, and Polosukhin}]{NIPS2017_3f5ee243}
Ashish Vaswani, Noam Shazeer, Niki Parmar, Jakob Uszkoreit, Llion Jones, Aidan~N Gomez, \L~ukasz Kaiser, and Illia Polosukhin. 2017.
\newblock \href {https://proceedings.neurips.cc/paper_files/paper/2017/file/3f5ee243547dee91fbd053c1c4a845aa-Paper.pdf} {Attention is all you need}.
\newblock In \emph{Advances in Neural Information Processing Systems}, volume~30. Curran Associates, Inc.

\bibitem[{Wei et~al.(2022)Wei, Wang, Schuurmans, Bosma, hsin Chi, Xia, Le, and Zhou}]{Wei2022ChainOT}
Jason Wei, Xuezhi Wang, Dale Schuurmans, Maarten Bosma, Ed~Huai hsin Chi, F.~Xia, Quoc Le, and Denny Zhou. 2022.
\newblock Chain of thought prompting elicits reasoning in large language models.
\newblock \emph{ArXiv}, abs/2201.11903.

\bibitem[{Wu et~al.(2023)Wu, Wang, Wan, Jiao, and Lyu}]{wu2023chatgpt}
Haoran Wu, Wenxuan Wang, Yuxuan Wan, Wenxiang Jiao, and Michael Lyu. 2023.
\newblock Chatgpt or grammarly? evaluating chatgpt on grammatical error correction benchmark.
\newblock \emph{arXiv preprint arXiv:2303.13648}.

\bibitem[{Yang et~al.(2023)Yang, Li, Zhang, Chen, and Cheng}]{yang2023exploring}
Xianjun Yang, Yan Li, Xinlu Zhang, Haifeng Chen, and Wei Cheng. 2023.
\newblock Exploring the limits of chatgpt for query or aspect-based text summarization.
\newblock \emph{arXiv preprint arXiv:2302.08081}.

\bibitem[{Yannakoudakis et~al.(2011)Yannakoudakis, Briscoe, and Medlock}]{yannakoudakis-etal-2011-new}
Helen Yannakoudakis, Ted Briscoe, and Ben Medlock. 2011.
\newblock \href {https://aclanthology.org/P11-1019} {A new dataset and method for automatically grading {ESOL} texts}.
\newblock In \emph{Proceedings of the 49th Annual Meeting of the Association for Computational Linguistics: Human Language Technologies}, pages 180--189, Portland, Oregon, USA. Association for Computational Linguistics.

\end{thebibliography}

\clearpage
\appendix

\begin{figure}[t!]
    \centering 
    \includegraphics[width=1\linewidth]{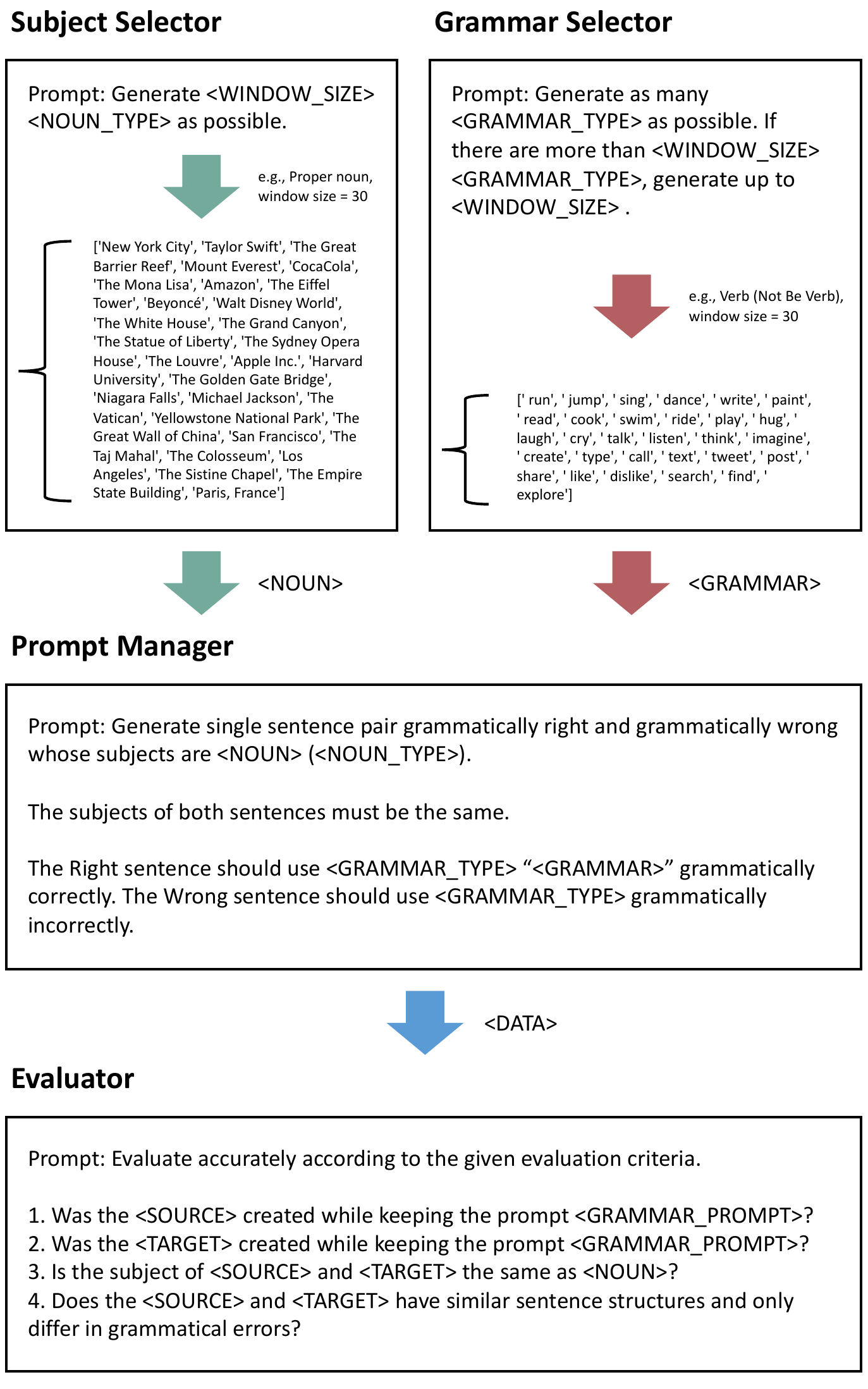}
    \caption{Schematic depiction of overall architecture}
    \label{Fig:architecture}
    \vspace{-1em}
\end{figure}




\section{Statistical Analysis: Error Type Distributions}
\label{sec:appendix_a}

Table \ref{Tab:Error_distribution} shows the ERRANT error type distributions of ChatLang-8 and The Other Corpora. We find that ChatLang-8 are most evenly distributed while the other corpora have highly unbalanced grammar labels.

\section{Architecture}
\label{sec:appendix_b}

As shown in Figure \ref{Fig:architecture}, our framework consists of Subject Selector, Grammar Selector, Prompt Manager, and Evaluator. Subject Selector and Grammar Selector ensure the diversity of the generated data, Prompt Manager maintains the format of the generated data, and Evaluator manages the quality of the generated data.

\begin{table*}[t]
\resizebox{\textwidth}{!}{%
\begin{tabular}{lc|cccccc}\toprule
  & ChatLang-8 & FCE & Lang-8 & NUCLE & W\&I+LOCNESS (Train) & W\&I+LOCNESS (Dev) & W\&I+LOCNESS (Test)\\ \midrule
ADJ & 1.44 & 1.36 & 1.25 & 1.58 & 1.52 & 1.48 & 1.05 \\
ADJ:FORM & 3.26 & 0.28 & 0.19 & 0.27 & 0.24 & 0.21 & 0.18 \\
ADV & 1.55 & 1.94 & 3.37 & 1.95 & 1.51 & 1.51 & 1.45\\
CONJ & 8.18 & 0.67 & 0.98 & 0.71 & 0.51 & 0.58 & 0.75\\
CONTR & 2.44 & 0.32 & 0.99 & 0.11 & 0.3 & 0.39 & 0.32\\
DET & 2.51 & 10.86 & 11.93 & 15.98 & 11.25 & 10.43 & 10.41 \\
MORTH & 3.73 & 1.9 & 1.62 & 3.14 & 1.85 & 2.07 & 2.5 \\
NOUN & 1.46 & 4.57 & 4.51 & 3.8 & 4.36 & 4.3 & 2.89 \\
NOUN:INFL & 3.69 & 0.5 & 0.18 & 0.12 & 0.12 & 0.13 & 0.28 \\
NOUN:NUM & 4.03 & 3.34 & 4.28 & 8.13 & 4.05 & 3.29 & 4.07 \\
NOUN:POSS & 3.83 & 0.51 & 0.35 & 0.61 & 0.6 & 0.87 & 0.93 \\
ORTH & 7.56 & 2.94 & 3.99 & 1.62 & 4.77 & 4.61 & 8.03 \\
OTHER & 0 & 13.26 & 26.62 & 25.65 & 12.76 & 12.84 & 15.69 \\
PART & 3.83 & 0.29 & 0.5 & 0.46 & 0.84 & 0.79 & 0.49 \\
PREP & 3.18 & 11.21 & 8 & 7.69 & 9.79 & 9.7 & 8.33 \\
PRON & 2 & 3.51 & 2.72 & 1.26 & 2.64 & 2.33 & 2.45 \\
PUNCT & 2.33 & 9.71 & 6.06 & 5.16 & 17.16 & 19.37 & 16.73 \\
SPELL & 4.03 & 9.59 & 4.45 & 0.26 & 3.74 & 5.07 & 4.63 \\
UNK & 0 & 3.13 & 0 & 2.57 & 2.59 & 2.24 & 1.41 \\
VERB & 6.54 & 7.01 & 6.52 & 4.31 & 5.86 & 5.27 & 5.09 \\
VERB:FORM & 5.23 & 3.55 & 2.56 & 3.49 & 3.56 & 3.09 & 3.1 \\
VERB:INFL & 12.07 & 0.19 & 0.15 & 0.01 & 0.04 & 0.07 & 0.12 \\
VERB:SVA & 0.95 & 1.52 & 1.58 & 3.47 & 2.23 & 1.94 & 2.28 \\
VERB:TENSE & 12.6 & 6.04 & 6.03 & 7.01 & 6.07 & 6.2 & 5.43 \\
WO & 3.58 & 1.82 & 1.18 & 0.66 & 1.64 & 1.25 & 1.4 \\ \hline \midrule
Total (\%) & 100 & 100 & 100 & 100 & 100 & 100 & 100 \\ \hline
\bottomrule 
\end{tabular}
}
\caption{\small{Distributions of the ChatLang-8 and the other datasets with respect to 25 grammatical errors.}}
\label{Tab:Error_distribution}
\end{table*}

\end{document}